\documentclass[letterpaper, review]{article}
\usepackage{iccc}

\usepackage{times}
\usepackage{helvet}
\usepackage{courier}
\usepackage[utf8]{inputenc} % allow utf-8 input
\usepackage[T1]{fontenc}    % use 8-bit T1 fonts
\usepackage[hidelinks]{hyperref}      % hyperlinks
\usepackage{url}            % simple URL typesetting
\usepackage{booktabs}       % professional-quality tables
\usepackage{amsfonts}       % blackboard math symbols
\usepackage{amsmath}
\usepackage{nicefrac}       % compact symbols for 1/2, etc.
\usepackage{microtype}      % microtypography
\usepackage{xcolor}         % colors
\usepackage{graphicx}
\usepackage{natbib}
\usepackage[bottom]{footmisc}

\graphicspath{ {./images/} }

\title{Controlling the image generation process with parametric activation functions}
\author{Ilia Pavlov\\
Creative Computing Institute, University of the Arts London, 
London UK\\
i.pavlov@arts.ac.uk\\
}
\begin{document}
\maketitle

\begin{abstract}
As image generative models continue to increase not only in their fidelity but also in their ubiquity, the development of tools that leverage direct interaction with their internal mechanisms in an interpretable way has received little attention. In this work, we introduce a system that allows users to develop a better understanding of the model through interaction and experimentation. By giving users the ability to replace activation functions of a generative network with parametric ones and a way to set the parameters of these functions, we introduce an alternative approach to control the network's output. We demonstrate the use of our method on StyleGAN2 and BigGAN networks trained on FFHQ and ImageNet, respectively.
\end{abstract}

\maketitle

\section{Introduction}

The field of Explainable AI (xAI) has always lagged behind the main body of machine learning research~\citep{10.1145/3588313}. This is also true for applications of generative artificial intelligence (AI)~\citep{Schneider2024ExplainableGA} in the field of computational creativity and it's use as a creativity support tool~\citep{Shneiderman2007CreativityST}.

This paper proposes an interactive tool that allows users to manipulate the behavior of a generative model by modifying its structure. Such interactions have the potential to expand the non-expert user's understanding of the underlying mechanisms behind the model's output and promote an increase in AI literacy. 

The tool allows users to replace static activation functions with parametric ones~\citep{Apicella2020ASO}. This gives users the ability to manually set the parameters of these functions, and thereby affect their outputs.

\section{Background}

\begin{figure*}[!h]
    \centering
    \includegraphics[width=\textwidth]{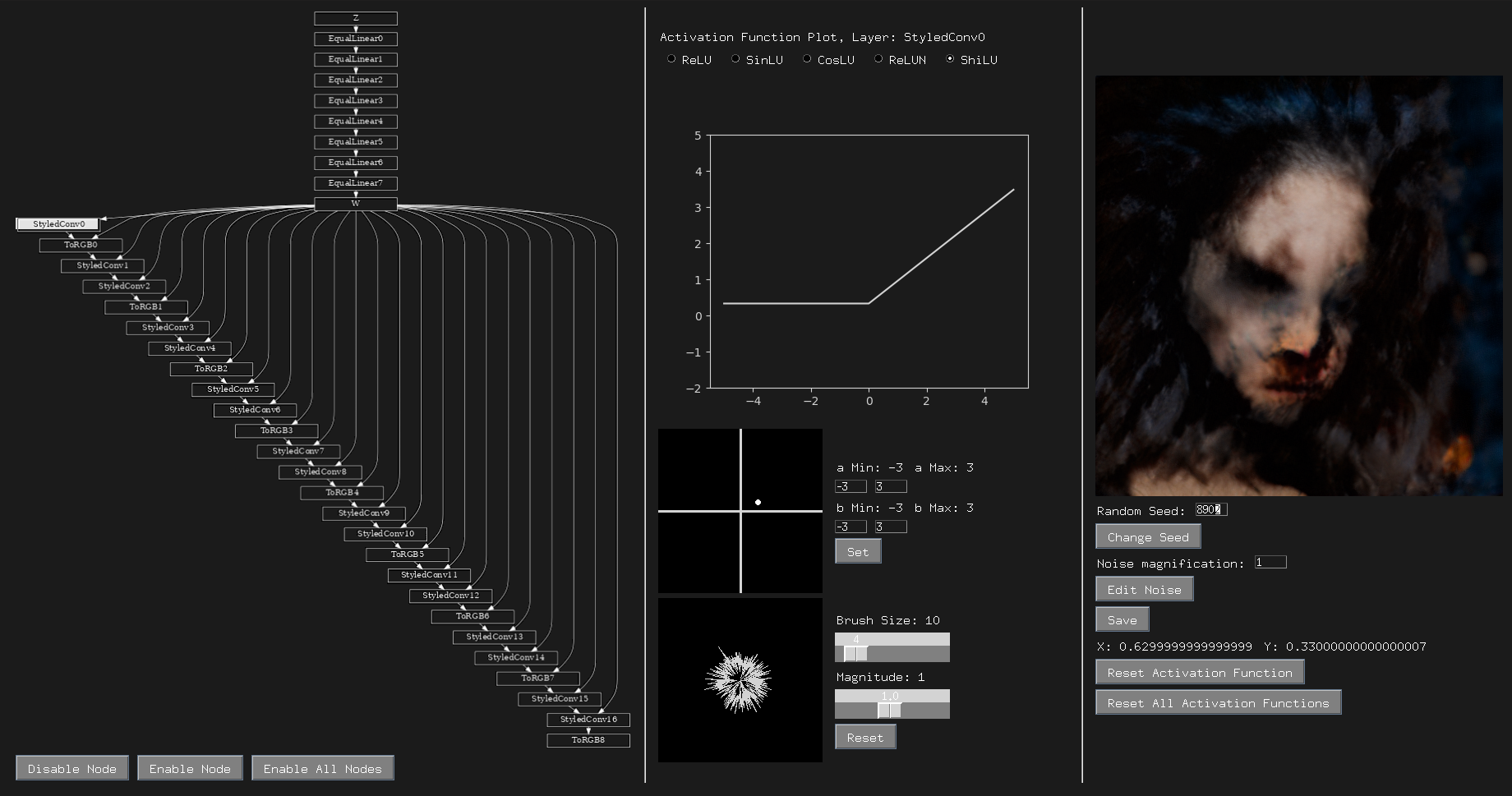}
    \caption{A Graphical User Interface was developed for our control method. The left side is responsible for picking neural layers, the center side is responsible for selecting activation functions and adjusting their parameters if applicable, and the right side allows for real-time result visualization.}
\end{figure*}

xAI as a field aims to develop methods to explain how specific outcomes are reached by an AI model~\citep{Weller2019TransparencyMA, Samek2017ExplainableAI}. Specific techniques for models working with visual data include Saliency Maps~\citep{Petsiuk2020BlackboxEO}, Class Activation Mapping~\citep{Zhou2015LearningDF}, Layer-Wise Relevance Propagation~\citep{Binder2016LayerWiseRP}.

Machine learning approach to generative models works by mapping a data point drawn from the latent distribution Z to a data point taken from the data distribution X~\citep{Ruthotto2021AnIT}. In practice, distribution Z is typically a Gaussian distribution, and distribution X is real-world or simulated data, such as images, audio, or text. There are different ways one can achieve this, some examples include the following archetypes: Generative Adversarial Networks (GANs)~\citep{Goodfellow2014GenerativeAN}, Variational Autoencoders~\citep{Kingma2013AutoEncodingVB}, and Diffusion models~\citep{Saharia2022PhotorealisticTD}. It is worth to note that generative models are not exclusive to neural networks and expand into evolutionary models~\citep{Cook2018RedesigningCC}, agent-based models~\citep{Delarosa2020GrowingMM} and others~\citep{Briot2019ComputationalCT}. 

The manipulation of the generative network’s feature maps is not a new area of research. The method described in the Network Bending paper~\citep{e24010028} uses deterministic transformations to affect the output of the network during inference. To make this more effective they introduce a procedure for clustering activation maps based on their spatial similarity. This allows for the grouping of feature maps, which correspond to different features of the generated image. 

PandA~\citep{Oldfield2022PandAUL} is a method for discovering factors responsible for image appearance in feature map space. This then enables users to edit the output images with pixel accuracy by modifying these factors.

The manipulation of the behaviour of generative neural networks can be also achieved by inserting small-scale neural networks in-between layers and training them for a short time, as shown in this paper~\citep{Aldegheri2023HackingGM}. This introduces deviations from original data distribution the network was trained on, while also maintaining an overall structure of images.

Another prominent method used to manipulate the generative network's output is latent space exploration~\citep{Klys2018LearningLS, Corneanu2024LatentPaintII, Shen2019InterpretingTL}. There is a large variety of work, detailing various approaches to the manipulation of network's latent space. For example, papers ~\citep{Bontrager2018DeepIE, Fernandes2020EvolutionaryLS} propose the use evolutionary algorithms for this task. 

\section{Control Method}
In this section, we introduce our GUI and examine the multiple parametric activation functions we experimented with. The code for this paper can be found on GitHub\footnote{\url{https://github.com/locsor/generativeControlUI}} and the video demonstration of it's use is available on YouTube.\footnote{\url{https://youtu.be/qkP9DHLicwM}}

Our graphical user interface, as shown in Figure 1, allows users to:
\begin{itemize}
  \item Select the desired activation functions (left).
  \item Replace selected activation functions with parametric ones (center-top).
  \item Adjust the parameters of injected activation functions (center-bottom).
  \item Have a real-time feedback (right).
\end{itemize}
The GUI also has features for editing the input latent vector (center-bottom) and disabling specific layers outright (left-bottom).

All these features provide control over the structure of the network and showcase how different changes affect the outcome.

\begin{figure*}[!h]
    \centering
    \includegraphics[width=0.95\textwidth]{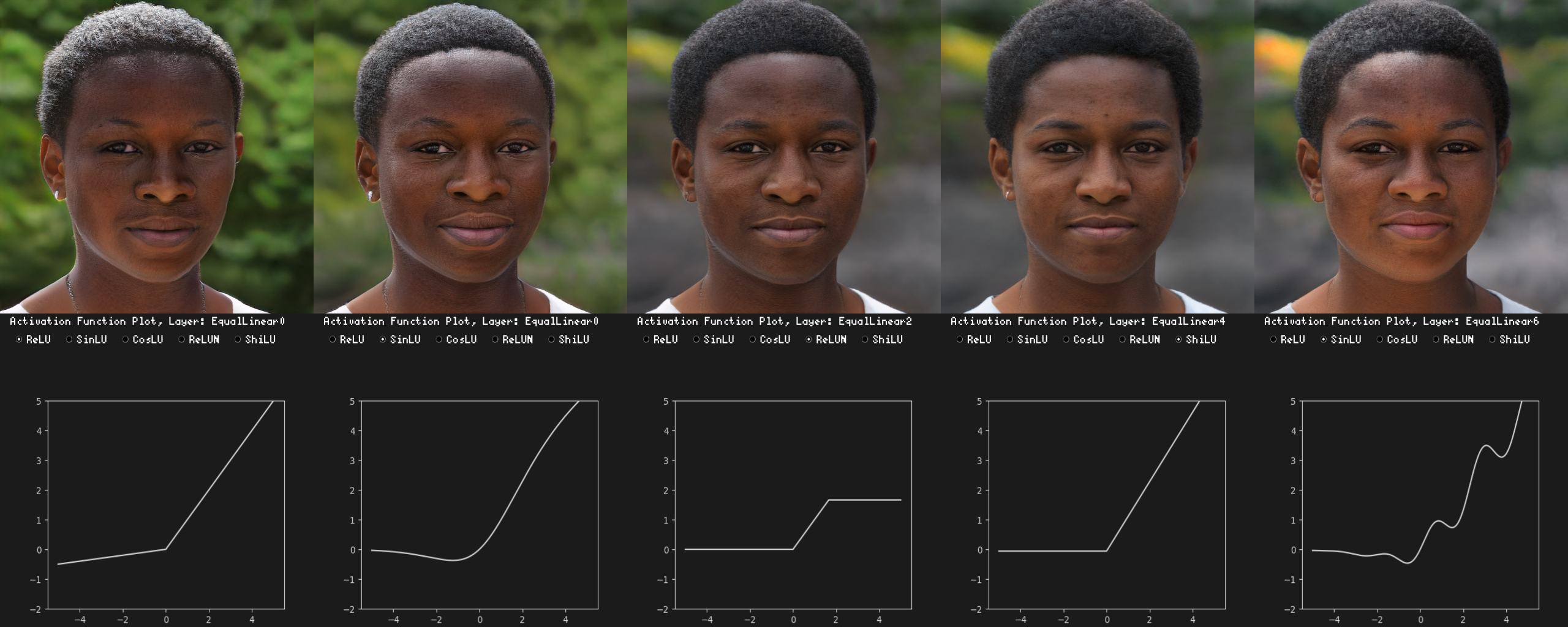}
    \includegraphics[width=0.95\textwidth]{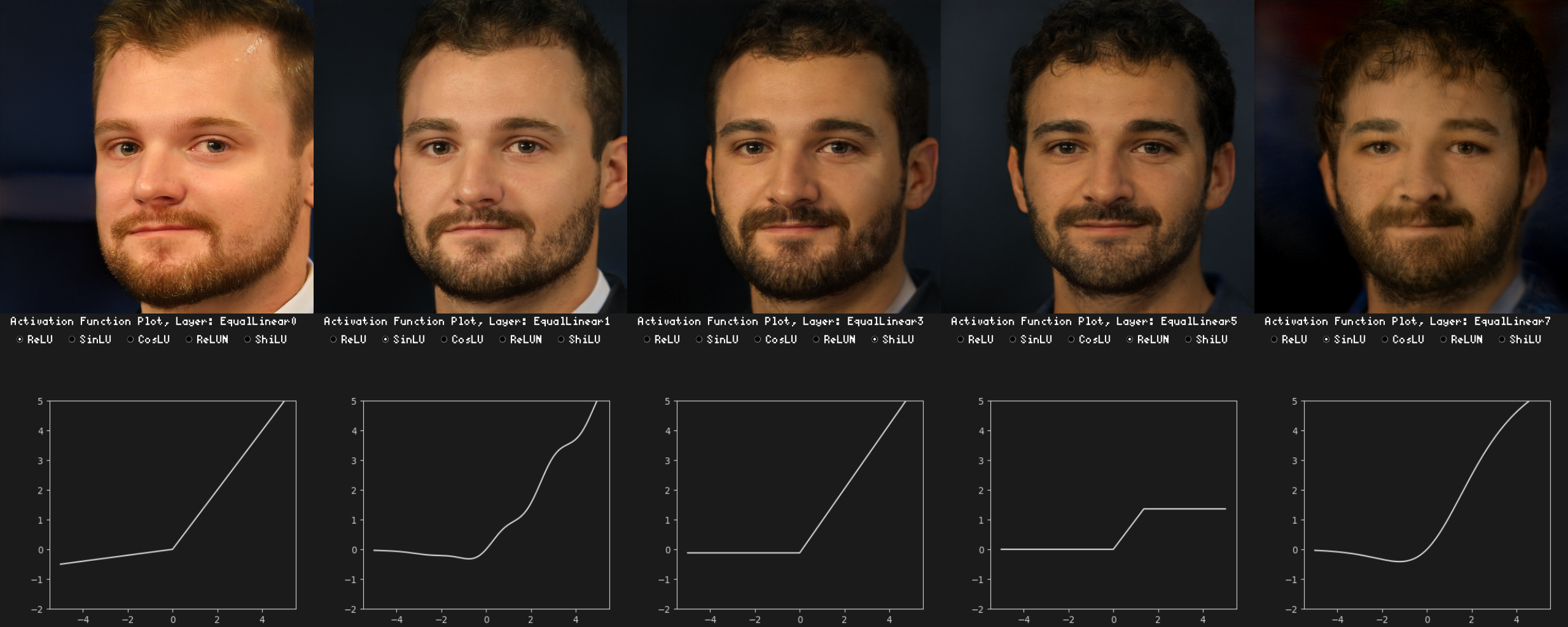}
    \caption{Examples of using parametric activation functions in the mapping network of the StyleGAN2. The left-most images in output rows (faces) are base outputs. Each image has one more base activation function replaced with a parametric one, compared with an image to its' left. Bottom rows show plots of parametric functions, which were used to produce the images above.}
\end{figure*}

\subsection{Activation functions}
The Sinu-Sigmoidal Linear Unit~\citep{math10030337} has two parameters, $a$ and $b$. The parameter $a$ is responsible for the amplitude of the sine function, while the parameter $b$ is responsible for the frequency of the sine function. Setting any of these parameters too high can lead to unpredictable results, which could be preferable for more abstract image generation. The full formula for the activation function is shown in (1), where $\sigma(x)$ is a sigmoid function.

\begin{equation}
SinLU(x) = (x+ a*sin(b*x)) * \sigma(x)
\end{equation}

Rectified Linear Unit N~\citep{202301.0463} or ReLUN is a modified version of the ReLU6 activation function with one parameter, which controls the angle of the function. The full formula for the activation function is shown in (2).

\begin{equation}
ReLUN(x) = min(max(0,x),n)
\end{equation}

Shifted Rectified Linear Unit~\citep{202301.0463} or ShiLU is a modified version of ReLU~\citep{relu} function that has two parameters $a$ and $b$. Parameter $a$ is responsible for the slope of the function, and parameter $b$ is responsible for shifting the function along the Y-axis. The full formula for the activation function is shown in (3).

\begin{equation}
ShiLU(x) = a*ReLU(x) + b
\end{equation}

Usage of such functions causes alterations in the output by changing the intermediate feature maps~\citep{Zeiler2013VisualizingAU} of the network during the inference process. Using these functions in multiple layers will cause a cascading effect of changes. This in turn can lead to a large amount of possible divergences from initial neural network state.

We also experimented with polynomial activation functions. Using polynomial functions as activation functions in neural networks has been proposed before, but never gained much popularity, due to the exploding gradient problem and increased complexity of training. The specific polynomial activation functions we used have been taken from a retracted paper~\citep{Gottemukkula}. The general formula is shown in (4).

\begin{equation}
\begin{gathered}
f_n(x) = \frac{\sum_{i=1}^{n} w_i \cdot \sigma(x)^i}{\sqrt{2}^n} \\
= \frac{w_0 \cdot \sigma(x)^0 + w_1 \cdot \sigma(x)^1 + \ldots + w_n \cdot \sigma(x)^n}{\sqrt{2}^n}
\end{gathered}
\end{equation}
Where each weight coefficient $w_i$ in this function is a potential control parameter in our system and $\sigma(x)$ is a sigmoid function.

\begin{figure*}[!ht]
    \centering
    \includegraphics[width=0.95\textwidth]{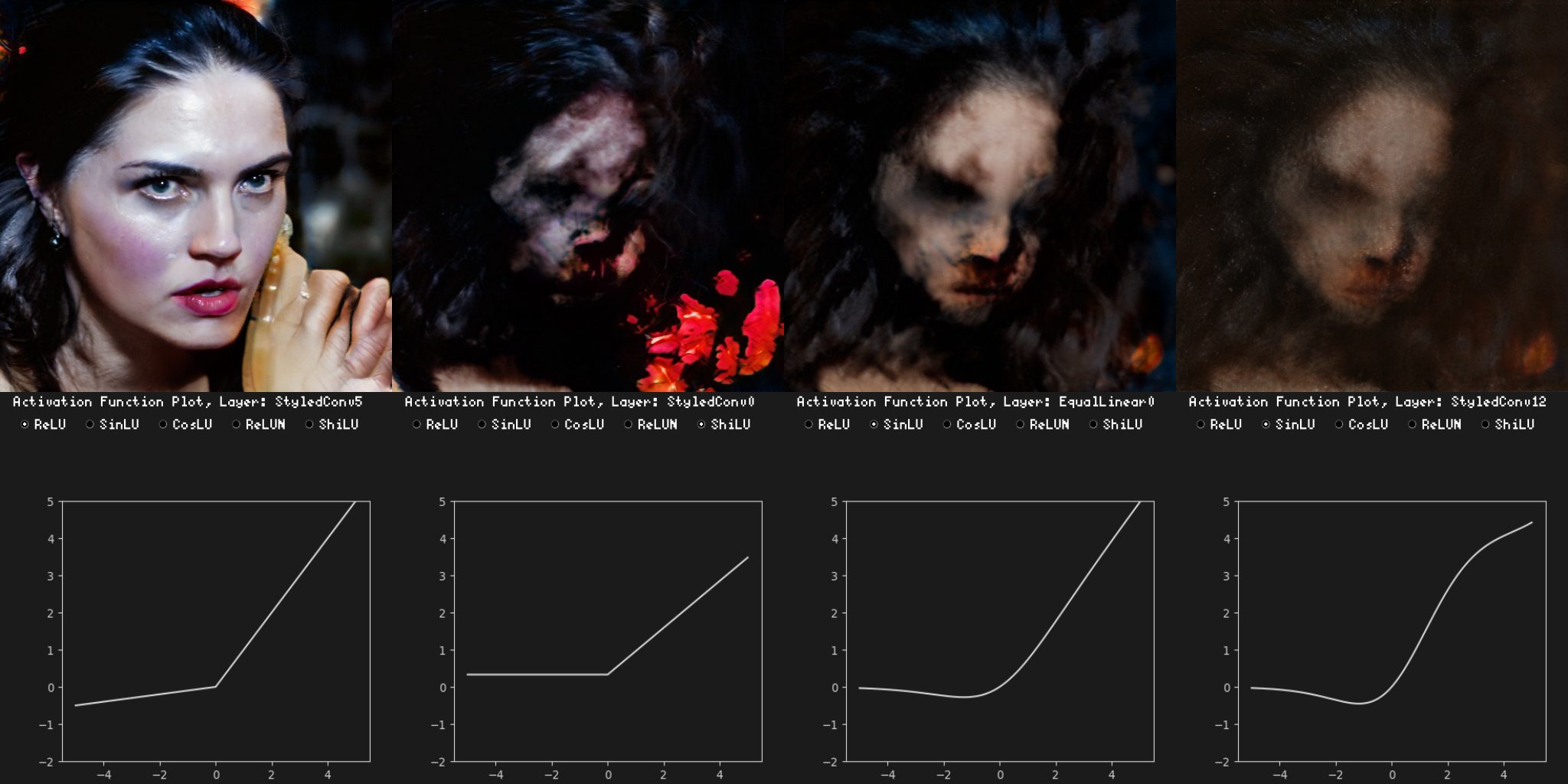}
    \includegraphics[width=0.95\textwidth]{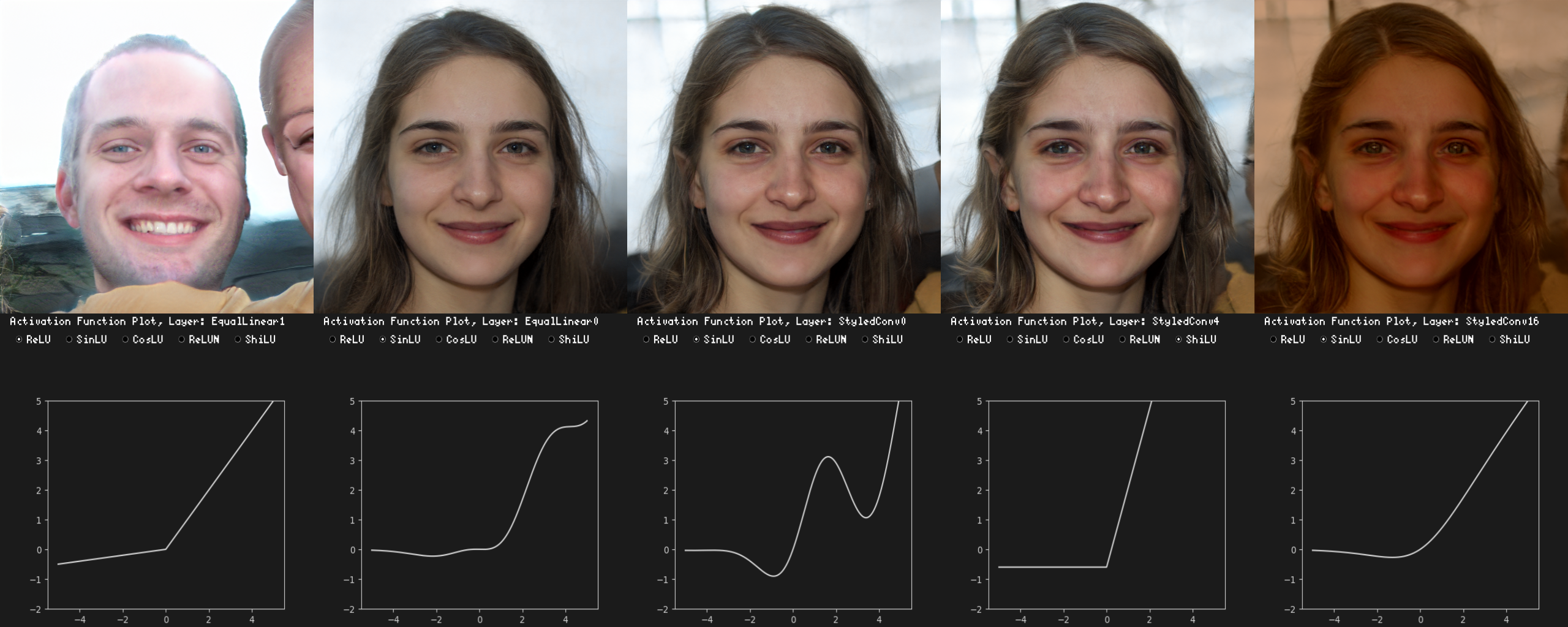}
    \caption{Examples of using parametric activation functions in both mapping and generator networks of the StyleGAN2. The left-most images in output rows (faces) are base outputs. Each image has one more base activation function replaced with a parametric one, compared with an image to its' left. Bottom rows show plots of parametric functions, which were used to produce the images above.}
\end{figure*}

\section{Experiments}

For base networks, we used the StyleGAN2~\citep{Karras2019AnalyzingAI} and BigGAN~\citep{Brock2018LargeSG} models. The choice was motivated by their ubiquity in generative neural network research, stability, and speed.

The architecture of StyleGAN2 is split into two parts: a mapping network and a generator network. The purpose of a mapping network is to disentangle the input latent vector. The generator network consists of a series of convolutional blocks, each outputting an image with progressively increasing resolution. To get the final image, these intermediate images are scaled up and summed.
Applying the parametric activation functions to StyleGAN2 had the following results: 

\begin{itemize}
    \item Using them in the mapping network affected the image structure. These structural changes are due to parametric activation functions affecting the resulting output of the mapping network, i.e. the disentangled latent vector.
    \item Earlier layers of the mapping network offered a high level of granularity with respect to changes in the image.
    \item Using them in the generator network affected both style and, in cases of earlier layers, the structure of images.
    \item Later layers of the generator network caused changes to images in a more basic manner, such as altering the overall coloration of the image.
\end{itemize}

As shown in our examples in Figures 2 and 3, the usage of controllable activation functions can be a viable way to control the image generation process. Slight changes to the parameters of these functions will lead to slight changes in the resulting image, while larger changes lead to more significant and unpredictable alterations to both the structure and style of the image. For example, a user can infer that the latter quality is not desirable for generating realistic images, it can allow for the generation of more abstract imagery.

\begin{figure*}[h]
    \centering
    \includegraphics[width=1\textwidth]{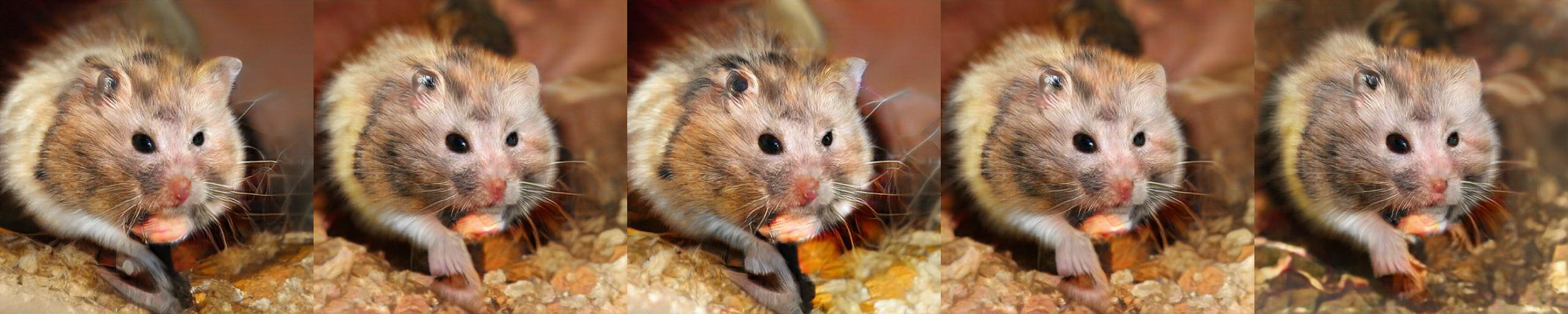}
    \caption{Output of a BigGAN model. The left-most image is an original, the rest were generated with SinLU or ReLUN applied to the second layer of the network. The parameter values were randomly chosen.}
\end{figure*}

\begin{figure*}[h]
    \centering
    \includegraphics[width=1\textwidth]{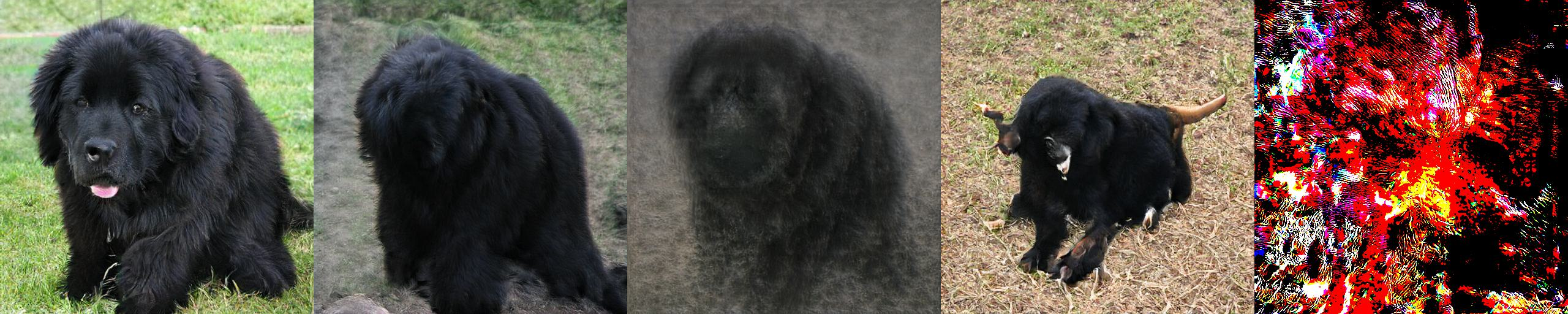}
    \includegraphics[width=1\textwidth]{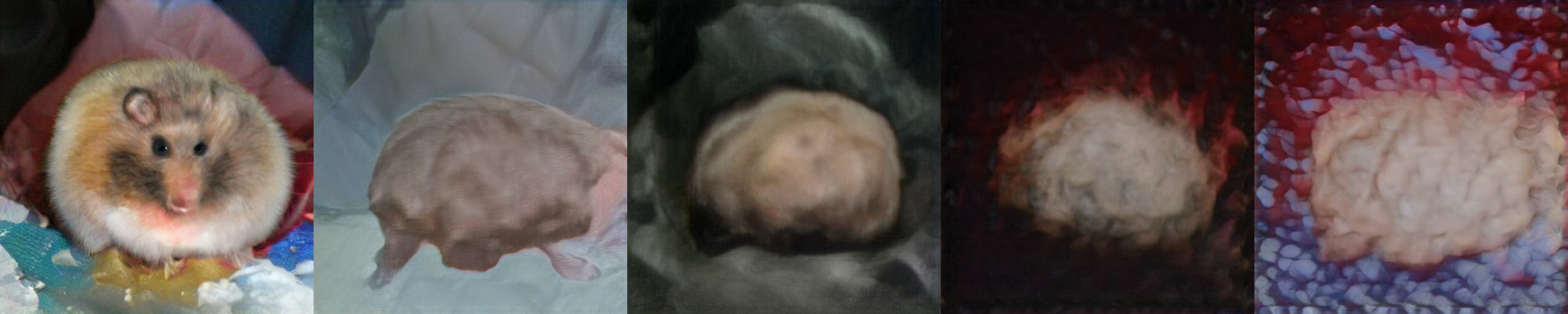}
    \caption{Output of a BigGAN model with 3rd order polynomial activation functions. The functions' parameters were kept within $\left[0.8, 1.2\right]$ interval. The left-most images are originals, the right-most are generated with 4 modified functions, and the rest were generated with only one modified function.}
\end{figure*}

BigGAN is a classic example of a GAN architecture but scaled up to improve both fidelity and variety of generated images. For our experiment, we used a class-conditional implementation of this network. While the changes in images are not as varied as those observed with StyleGAN2, they still demonstrate the ability to alter the content of the image to some extent. Examples of images generated using BigGAN are shown in Figures 4 and 5.

Polynomial functions were not as encouraging. Using them in more than 2 neural layers yielded non-ideal results, and they were extremely sensitive to small changes in parameter values. Examples of images generated by the BigGAN model modified with a 3rd degree polynomial activation functions are shown in Figure 5. We limited the parameter values to $\left[0.5, 1.5\right]$.

Overall, in a standard use case, a user will gradually modify the network and carefully select the values of the control parameters to achieve a satisfactory result. Since our method does not offer any way to single out any specific feature of the image, the process largely consists of trial-and-error experimentation. Through continued use and real-time visual feedback, users can learn how the structural state of the network influences the output of the generation process. This process is mainly aimed at incidental exploration, instead of a guided image generation.

\section{Conclusion}
In summary, this paper introduces a novel approach to control image generative networks, with the help of parametric activation functions, and examines how this tool can be used to teach non-expert users the importance of its various parts. But to confirm these speculations, a user study will need to be conducted.

The main limitation of the current implementation is, that unguided control over image generation is imprecise and limiting. While applying this method to the text-to-image generative network can alleviate this problem, there are still no guarantees that users will be able to find a satisfactory collection of activation functions and their parameters.

\bibliographystyle{iccc}
\bibliography{neur}

\end{document}